\pdfoutput=1

\documentclass[11pt]{article}

\usepackage[final]{acl}
\usepackage{listings}

\usepackage{times}
\usepackage{latexsym}

\usepackage[T1]{fontenc}

\usepackage[utf8]{inputenc}

\usepackage{microtype}

\usepackage{inconsolata}

\usepackage{graphicx}

\usepackage{tabularx}
\usepackage{booktabs}
\usepackage{siunitx}

\usepackage{enumitem}
\usepackage{xcolor}
\usepackage{tcolorbox}
\usepackage{colortbl}
\newcommand{\negdelta}[1]{\cellcolor{red!20}{#1}}
\newcommand{\posdelta}[1]{\cellcolor{green!20}{#1}}
\newcommand{\highposdelta}[1]{\cellcolor{green!40}{#1}}

\definecolor{lightyellow}{HTML}{fff2cc}
\definecolor{lightblue}{HTML}{cfe2f3}
\newcommand{\reducedstrut}{\vrule width 0pt height .9\ht\strutbox depth .9\dp\strutbox\relax}
\newcommand{\highlightyellow}[1]{%
  \begingroup
  \setlength{\fboxsep}{0pt}%
  \colorbox{lightyellow}{\reducedstrut#1}%
  \endgroup
}
\newcommand{\highlightblue}[1]{%
  \begingroup
  \setlength{\fboxsep}{0pt}%
  \colorbox{lightblue}{\reducedstrut#1}%
  \endgroup
}

\usepackage{hyperref}

\usepackage{cleveref}
\crefformat{section}{\S#2#1#3} 
\crefformat{subsection}{\S#2#1#3}
\crefformat{subsubsection}{\S#2#1#3}

\usepackage{tikz}
\usepackage[inkscapelatex=false]{svg}

\title{\textsc{Tulun}: Transparent and Adaptable Low-resource Machine Translation}

\author{Raphaël Merx$^\lambda$ \hspace{0.5cm} Hanna Suominen$^\psi$ \hspace{0.5cm} Lois Hong$^\zeta$ \\
{\bf Nick Thieberger$^\lambda$} \hspace{0.5cm} {\bf Trevor Cohn$^{\lambda}$} \hspace{0.5cm} {\bf Ekaterina Vylomova$^\lambda$} \\
$^\lambda$The University of Melbourne  \hspace{0.5cm} $^\psi$The Australian National University \\
$^\psi$University of Turku  \hspace{0.5cm} $^\zeta$Maluk Timor
}

\begin{document}
\maketitle
\begin{abstract}
Machine translation (MT) systems that support low-resource languages often struggle on specialized domains. While researchers have proposed various techniques for domain adaptation, these approaches typically require model fine-tuning, making them impractical for non-technical users and small organizations.
To address this gap, we propose \textsc{Tulun},\footnote{Tulun means "assistance" in Tetun, highlighting a philosophy of augmenting rather than replacing human expertise.} a versatile solution for terminology-aware translation, combining neural MT with large language model (LLM)-based post-editing guided by existing glossaries and translation memories.
Our open-source web-based platform enables users to easily create, edit, and leverage terminology resources, fostering a collaborative human-machine translation process that respects and incorporates domain expertise while increasing MT accuracy.
Evaluations show effectiveness in both real-world and benchmark scenarios: on medical and disaster relief translation tasks for Tetun and Bislama, our system achieves improvements of 16.90-22.41 ChrF++ points over baseline MT systems. Across six low-resource languages on the FLORES dataset, \textsc{Tulun} outperforms both standalone MT and LLM approaches, achieving an average improvement of 2.8 ChrF points over NLLB-54B. \textsc{Tulun} is publicly accessible at \href{https://bislama-trans.rapha.dev/}{bislama-trans.rapha.dev}.
\end{abstract}

\section{Introduction}

Machine translation (MT) systems have transformed how organizations manage their translation needs \cite{stefaniak_machine_2022, utunen_scale_2023}, yet domain accuracy and consistency remain a significant challenge, particularly for low-resource languages \cite{haddow_survey_2022, khiu_predicting_2024, marashian_priest_2025}. For instance, a health organization we work with in Timor-Leste struggled to leverage MT to accurately translate medical education materials from English to Tetun, despite having a glossary and a corpus of past translations that could inform MT output. Tetun, a low-resource language that is the lingua franca in Timor-Leste, lacks available corpora in the health domain \cite{merx_low-resource_2024}, making in-domain resources particularly valuable to improve MT accuracy.
This case exemplifies a broader challenge: model adaptation and deployment requires technical expertise, and commercial MT providers rarely offer low-resource language support, let alone terminology customization, leaving no practical option for a small organization to rely on MT for low-resource in-domain~translation.

\begin{figure}
    \centering
    \includesvg[width=1.0\linewidth]{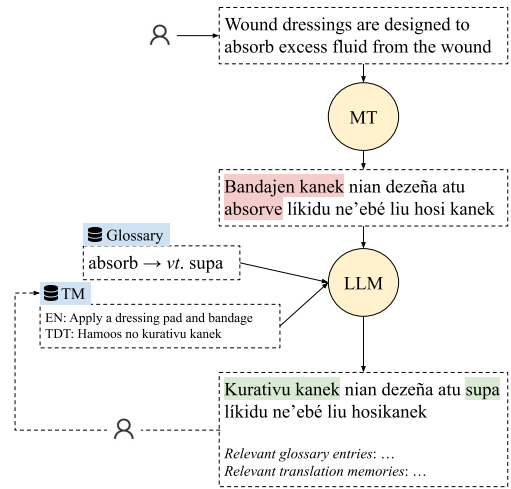}
    \caption{System overview with example translation from English to Tetun (en-tdt). The \highlightyellow{system components} and \highlightblue{data} are configurable by end-users.}
    \label{fig:system-architecture}
\end{figure}

Translation memories and terminology management are well-established tools in professional translation software, improving translation accuracy while reducing cognitive load \cite{dillon_translators_2006, drugan_translation_2023}. Research has demonstrated that lexicons can bring substantial accuracy gains, particularly for low-resource MT \cite{jones_gatitos_2023}. However, existing approaches to incorporate terminology constraints into neural MT systems typically require model fine-tuning \cite{reid-artetxe-2022-paradise, niehues-2021-continuous}, making them inaccessible to small organizations \cite{bane-etal-2023-coming}. Recent advances in large language models (LLMs) offer a promising alternative: while LLMs may underperform specialized MT systems for low-resource languages \cite{robinson-etal-2023-chatgpt}, their ability to adapt to new contexts at inference time \cite{brown_language_2020} makes them particularly suitable for terminology-aware post-editing \cite{raunak_leveraging_2023}.

To address these challenges, we propose \textsc{Tulun}, a versatile solution that combines neural MT with LLM-based post-editing, guided by existing glossaries and translation memories (Figure~\ref{fig:system-architecture}). \textsc{Tulun} continuously adapts as new entries are added to the translation memory and glossary. Packaged as an open-source\footnote{Code: \href{https://github.com/raphaelmerx/tulun/}{github.com/raphaelmerx/tulun/}, MIT license} web platform, it relies on a modular architecture that allows users to configure their choice of MT system, LLM, and retrieval options, as well as create, edit, and rely on terminology resources. To see a demo video of \textsc{Tulun}, visit \href{https://youtu.be/fQFwOxzR4MI}{https://youtu.be/fQFwOxzR4MI}.

Our system has the following characteristics:
\begin{itemize}[nosep]
    \item \textbf{Accurate}: On real-world medical and disaster relief translation tasks for Tetun and Bislama (national language of Vanuatu), our system shows impressive improvements of 16.90-22.41 ChrF++ points over baseline MT systems (\cref{sec:applied-scenarios}). A broader evaluation across six low-resource languages on the FLORES dataset shows \textsc{Tulun} outperforms both standalone MT and LLM approaches (\cref{sec:flores-eval}).
    \item \textbf{User-friendly}: Our usability study, based on the System Usability Scale (SUS), averages an excellent score of 81.25, with users rating the system's overall usefulness at 5/5 for their translation tasks (\cref{sec:human-eval}).
    \item \textbf{Adaptable}: Target language, MT model, and prompt are all configurable from the user interface (UI). Glossary and translation memories can be bulk-imported and managed through the UI (\cref{sec:system-design}).
    \item \textbf{Transparent}: Users can verify how their glossary entries and past translations inform the current translation.
    \item \textbf{Lightweight}: Easy to deploy (\cref{sec:deployment}), does not require model training.
\end{itemize}

Fundamentally, \textsc{Tulun} represents a shift in MT philosophy, moving away from the paradigm of users as passive consumers of opaque systems \cite{liebling-etal-2022-opportunities}, toward one where users' expertise and preferences actively shape the translation process \cite{liu_new_2025}. By making glossary and translation memory matches explicit to users, and by allowing configuration of the underlying data and systems, \textsc{Tulun} aims to foster a transparent, collaborative process that respects and leverages users' domain knowledge. This approach benefits low-resource in-domain translation, where local expertise is often the most valuable resource for producing accurate, culturally appropriate translations \cite{nekoto-etal-2020-participatory}.

\section{Related Work}

\paragraph{Glossary and translation memory integration in MT}
The integration of custom terminology and translation memories into MT systems can deliver more consistent, domain-adapted translations \cite{scansani-dugast-2021-glossary}.
Recent research has demonstrated that such lexical customization brings substantial accuracy gains, particularly for low-resource MT \cite{jones_gatitos_2023}. Approaches to incorporate terminology constraints into neural MT models include replacing source words with their target translation in the source \cite{reid-artetxe-2022-paradise}, and prepending dictionary entries to the source text \cite{niehues-2021-continuous}. However, these approaches typically require either custom models, or model fine-tuning, which can be resource-intensive for smaller organizations \cite{bane-etal-2023-coming}, and prone to catastrophic forgetting \cite{saunders_domain_2022}. \textsc{Tulun} addresses this gap by providing a deployable solution that requires no model training while delivering terminology-consistent~translations.


\paragraph{LLMs and Automated Post-Editing (APE)}
The ability for LLMs to adapt to new tasks at inference time \cite{brown_language_2020} makes them of interest to both MT \cite{moslem_adaptive_2023} and related tasks, such as synthetic data generation and automated post-editing \cite{moslem_domain_2023}. While their MT accuracy can lag behind that of specialized MT models when translating into low-resource languages \cite{robinson-etal-2023-chatgpt} or in specialized domains \cite{uguet-etal-2024-llms}, \citet{raunak_leveraging_2023} find that combining specialized MT with an LLM for APE results in more accurate translations than each module used in isolation (measured using COMET on high-resource language pairs). Confirming the potential of LLMs at the post-editing stage, \citet{ki-carpuat-2024-guiding} give external feedback to an LLM to improve MT outputs, and \citet{lu-etal-2025-mqm} find that LLMs can identify and correct translation mistakes across high and low-resource language pairs. These findings suggest that LLMs can be valuable for terminology-aware post-editing, where their adaptation capabilities are combined with the robustness of specialized MT systems.

\paragraph{Our contribution}
Building on research showing the potential of terminology-aware translation and LLM-based post-editing, \textsc{Tulun}  extends the applicability of these techniques through a modular user-friendly interface, and demonstrates their effectiveness across diverse scenarios, from applied use cases (\cref{sec:applied-scenarios}) to systematic evaluation (\cref{sec:flores-eval}). Our system serves as both a practical tool for immediate use, and as a research platform that demonstrates how these models can be effectively combined. To our knowledge, it is the only open-source terminology-aware MT tool that supports low-resource languages like Tetun and Bislama.

\section{System Design \& Implementation}

\begin{figure}
    \centering
    \includegraphics[width=\linewidth]{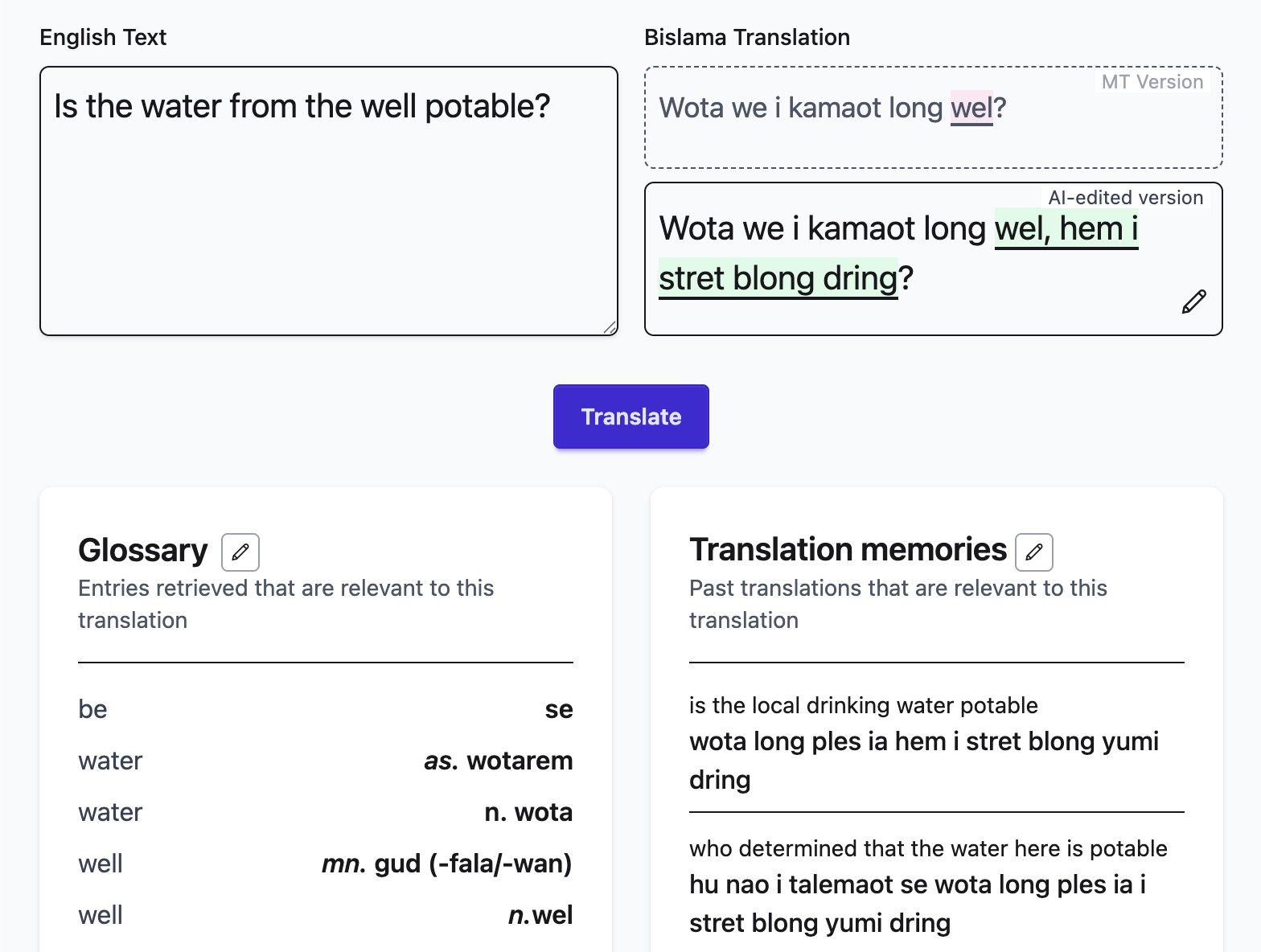}
    \caption{Translation View with the MT text, post-edited text, and the glossary entries and past translations relevant to this translation.}
    \label{fig:translation-view}
\end{figure}

\subsection{System Design}
\label{sec:system-design}

\paragraph{Translation View}
\label{para:translation-view}
When users open \textsc{Tulun}, they are presented with the Translation View, where they can enter a sentence or paragraph, and have it first machine translated, then post-edited using an LLM (Figure~\ref{fig:translation-view}). Post-editing changes are highlighted in the machine translated text (in red) and in the post-edited final translation (in green). In addition, users are presented with the relevant glossary entries and similar sentences retrieved to guide the LLM for post-editing. For example, in Figure~\ref{fig:translation-view}, ``potable'' is translated incorrectly by the MT model, but the LLM identifies the correct translation (``stret blong dring'') from the translation memory, and applies this change at the post-editing phase.

\paragraph{Glossary and Translation Memory View}
Both the glossary entries and the translation memories are editable by end-users (if they are given permission to do so, a setting configured through the admin). This allows users to iteratively improve the translation quality, by adding or correcting entries as missing or incorrect entries are found. In addition, data can be bulk-imported from a CSV, and a new translation memory can be added from the current (source, final translation) pair directly from the Translation View in a dedicated modal.


\paragraph{System Configuration}
\label{sec:sys-config}
Admin users can set through the web UI: (1) Site metadata, including target language and site title, (2) MT model, with a choice between Google Translate or any model available on HuggingFace through its ``translation'' pipeline,\footnote{\href{https://huggingface.co/models?pipeline_tag=translation}{huggingface.co/models?pipeline\_tag=translation}} and (3) LLM configuration for post-editing, including the choice of LLM among the hundreds of providers supported by LiteLLM,\footnote{\href{https://docs.litellm.ai/}{docs.litellm.ai}} the system prompt, and the number of translation memories retrieved for in-context~learning.

\begin{figure}
    \centering
    \includegraphics[width=\linewidth]{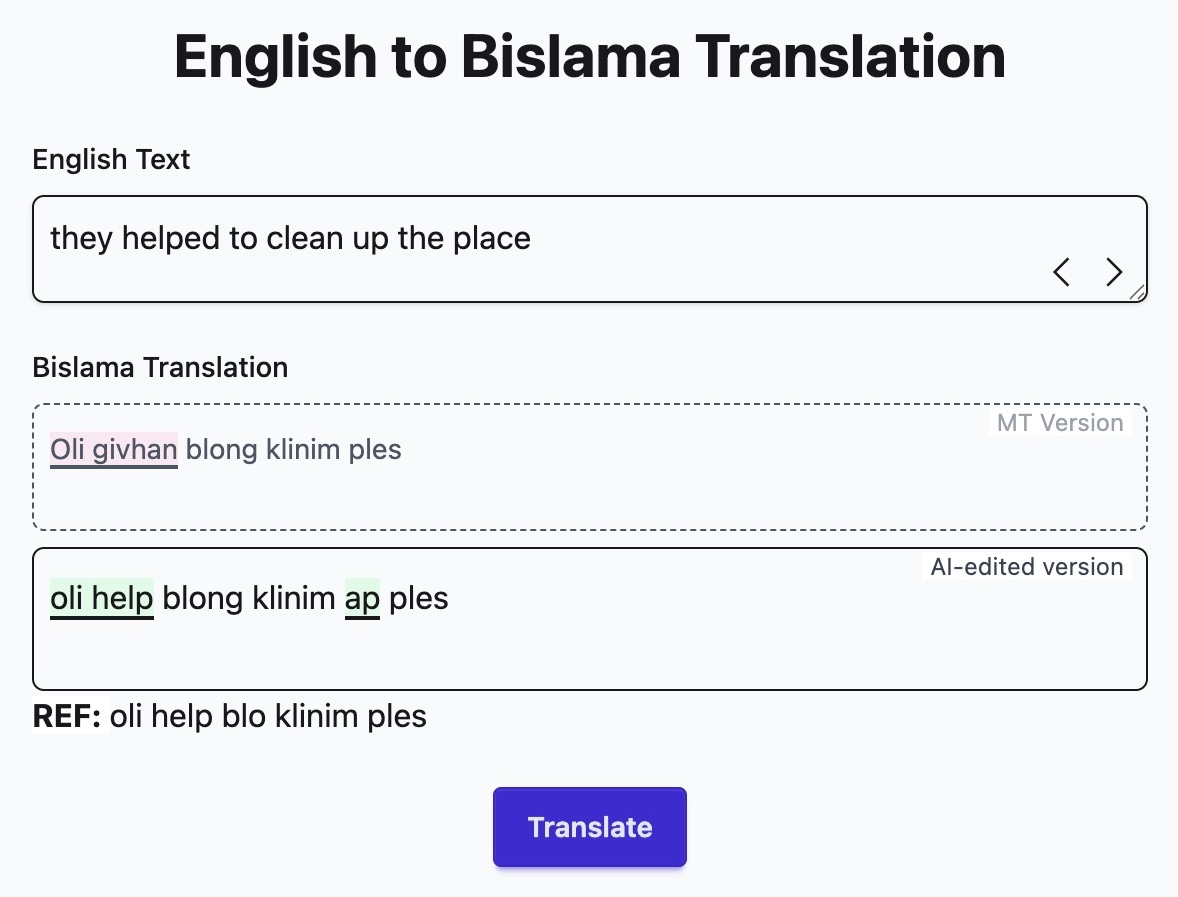}
    \caption{Eval mode: users can browse evaluation results, and see the reference translation}
    \label{fig:eval-mode}
\end{figure}

\paragraph{Evaluation Mode}
\textsc{Tulun} includes a dedicated evaluation feature that allows users to assess translation quality against reference translations. After uploading an evaluation dataset through the admin UI, users can navigate through these test translations within the Translation View (Figure~\ref{fig:eval-mode}). When a source sentence from the evaluation set is entered, the system automatically displays both the system-generated translation and the human reference for comparison. This helps users identify areas where improvements to the glossary, translation memory, LLM system prompt, or MT model selection might be beneficial.

\subsection{System Architecture}

\paragraph{Backend}
We implement \textsc{Tulun} as a configurable Django project, with data models for the glossary and translation memory, a \texttt{Translator} class that implements compatibility with either Google Translate (through the Cloud Translation API)\footnote{\href{https://pypi.org/project/google-cloud-translate/}{pypi.org/project/google-cloud-translate/}} or the HuggingFace \href{https://huggingface.co/docs/transformers/en/main_classes/pipelines\#transformers.TranslationPipeline}{Translation pipeline}, and an LLM post-editing layer that supports hundreds of providers via LiteLLM, with the choice of model and prompt configurable through the web UI.

\paragraph{Glossary and Translation Memory Retrieval}
At the post-edition stage, relevant glossary entries are retrieved using \{1,2\}-gram overlap with the input text tokens (tokenization is handled by spaCy's \texttt{en\_core\_web\_sm} model). Relevant translation memories are the top $N$ (where $N$ is configurable) BM25 matches between the input text and the source side of the memory, implemented using the Tantivy library.\footnote{\href{https://pypi.org/project/tantivy/}{pypi.org/project/tantivy/}} We select BM25 for retrieval because of its high performance on retrieving translation memories for MT through in-context learning \cite{bouthors_retrieving_2024}.

\paragraph{Prompt Design}
The glossary and translation memories are injected in the LLM prompt to inform post-editing (see an example prompt in Appendix~\ref{sec:example-prompt}). For all evaluations in Section~\ref{sec:evaluation}, we rely on a system prompt that includes few-shot examples \cite{brown_language_2020} with chain of thought reasoning \cite{wei_chain--thought_2022}. The prompt can be manually adjusted in the admin UI.

\paragraph{Deployment}
\label{sec:deployment}
We package \textsc{Tulun} using Docker and Docker Compose, allowing organizations to run the system on their infrastructure with minimal setup. The Docker configuration handles dependencies and environment configuration, while Docker Compose simplifies the orchestration process. This packaging approach ensures that the system can be deployed consistently across different~environments.

\section{Evaluation}
\label{sec:evaluation}

\subsection{Applications: Tetun Medical Translation and Bislama Disaster Relief Translation}
\label{sec:applied-scenarios}

We evaluate \textsc{Tulun} in two real-world low-resource language settings with distinct domain needs. For Tetun medical translation, we collaborate with Maluk Timor,\footnote{\href{https://maluktimor.org/}{maluktimor.org}} a health organization in Timor-Leste that regularly translates health education materials from English to Tetun. This translation work is needed as health workers (particularly nurses and community health workers) are most comfortable learning in Tetun rather than English or Portuguese \cite{greksakova_tetun_2018}. Maluk Timor reports that professional translation costs represent a significant organizational expense, and while they utilize machine translation, MT outputs typically require substantial post-editing to ensure accuracy and domain-appropriateness. For Bislama disaster relief translation, we partner with researchers working on a Pacific Creoles project\footnote{\href{https://researchportalplus.anu.edu.au/en/projects/modelling-pacific-creole-languages}{anu.edu.au/projects/modelling-pacific-creole-languages}} who need to translate transcripts while maintaining consistent terminology. Both scenarios provide practical test cases for \textsc{Tulun}'s ability to support organizations working with specialized domains in low-resource languages.

\subsubsection{MT Accuracy Evaluation}

\paragraph{Problem Statement}
From both organizations, we get a glossary (Tetun medical glossary: 2,698 entries; Bislama dictionary: 5,769 entries) and a translation memory (1,018 sentences for Tetun, 3,353 utterances for Bislama). We reserve some of the translation memory for evaluation (451 sentences for Tetun, 841 utterances for Bislama). Both datasets belong to their respective organizations, but are available upon request for research purposes with appropriate data sharing agreements.

\paragraph{Choice of Baseline and Prompt}
Given that neither Tetun nor Bislama are part of NLLB, we initially use MADLAD-400 10B \cite{kudugunta_madlad-400_2023} as baseline. We find that it performs poorly on Bislama, often copying the English source, and choose to also evaluate OPUS-MT models as an alternative baseline\footnote{\href{https://huggingface.co/Helsinki-NLP/opus-mt-en-tdt}{opus-mt-en-tdt}; \href{https://huggingface.co/Helsinki-NLP/opus-mt-en-bi}{opus-mt-en-bi}} \cite{tiedemann_democratizing_2024}. For post-editing, we use Gemini 2.0 Flash \cite{gemini_team_gemini_2024,gemini_team_introducing_2024}, with a prompt that describes the post-editing task and gives a few examples (see an example in Appendix~\ref{sec:example-prompt}).

\begin{table}
\small
\begin{tabularx}{\linewidth}{XSSS}
\toprule
Method & {\textsc{tdt}} & {\textsc{bis}} & {\textsc{\textbf{avg}}} \\
\midrule
\multicolumn{4}{l}{\textit{NMT only}} \\
MADLAD-400-10B & 35.78 & 15.22 & 24.37 \\
opus-mt-en-** & 16.01 & 32.60 & 24.31 \\
\midrule
\multicolumn{4}{l}{\textit{LLM only}} \\
Gemini, 0-shot & 44.05 & 37.78 & 39.63 \\
Gemini, 10-shot & 44.59 & 45.37 & 44.98 \\
\midrule
\multicolumn{4}{l}{\textit{Ours: NMT + LLM APE}} \\
MADLAD + Gemini & 47.87 & 47.94 & 47.91 \\ 
\quad $\Delta$ vs MADLAD & \highposdelta{+12.09} & \highposdelta{+32.72} & \highposdelta{+22.41} \\
opus-mt + Gemini & 34.27 & 48.14 & 41.21 \\
\quad $\Delta$ vs opus-mt & \highposdelta{+18.26} & \highposdelta{+15.54} & \highposdelta{+16.90} \\
\bottomrule
\end{tabularx}
\caption{ChrF++ score comparison on test sets for Tetun (tdt) and Bislama (bis). LLM only uses the system prompt from \cite{caswell_smol_2025}, and examples from the Tetun/Bislama corpus.}
\label{tab:tdt-bis}
\end{table}


\paragraph{Results}
Our approach demonstrates substantial translation quality improvements over baseline MT systems for both settings, with LLM post-editing yielding ChrF++ gains of 16.90-22.41 points (Table~\ref{tab:tdt-bis}). Qualitatively, we observe that for both settings, LLM post-editing helps (1) improve in-domain terminology translation (see an example in Figure~\ref{fig:translation-view}) and (2) repair hallucinations that are frequent for out-of-domain MT inference \cite{raunak-etal-2021-curious}, with the latter particularly relevant for the speech domain covered in our Bislama experiment.

\subsubsection{Usability and Usefulness Study}
\label{sec:human-eval}


\paragraph{Usability}
We perform a usability study of the \textsc{Tulun} interface using the System Usability Scale (SUS, \citeauthor{brooke_sus_1996} \citeyear{brooke_sus_1996}). We collect two responses, one from the clinical director at Maluk Timor, and the other from a linguist working with Bislama. We get an average SUS score of $81.25$, corresponding to an excellent perceived usability \cite{bangor_empirical_2008}.

\paragraph{Usefulness}
To measure usefulness, we adapt the Technology Acceptance Model (TAM, \citeauthor{venkatesh_user_2003} \citeyear{venkatesh_user_2003}) questions on general usefulness to our translation context. We get average scores between 4 and 5 for all questions (out of 5), with a 5/5 score for overall usefulness (``Overall, I find this system useful for my translation tasks''), a 4.5/5 score for the system impact on translation quality (``Using this system improves the quality of my translations''), and a 4.5/5 score for the system's helpfulness to translate technical content (``Using this system makes it easier to translate technical/specialized content'').

We report all questions, with scores for each annotator, in Appendix~\ref{app:usability-usefulness}.

\subsection{Generalizable Evaluation: FLORES-200}
\label{sec:flores-eval}

\paragraph{Languages}
To measure the broader efficacy of our solution, we work with six low-resource languages (Tok Pisin \textsc{tpi}, Dzongkha \textsc{dzo}, Quechua \textsc{quy}, Rundi \textsc{run}, Lingala \textsc{lin}, Assamese \textsc{asm}), spanning four continents and three different scripts. We select these languages because they are all (1)~low-resource (2) institutionalized, which makes them more likely to be standardized and in demand for MT \cite{bird_must_2024}, (3) part of the FLORES-200 evaluation benchmark \cite{costa-jussa_scaling_2024} and (4) represented in the \textsc{Gatitos} glossary project \cite{jones_gatitos_2023}.

\paragraph{Data}
We evaluate on all 1,012 sentences from the FLORES-200 "devtest" split, using NLLB-54B as a baseline MT model.\footnote{We get FLORES-200 translations by NLLB-54B from \href{https://tinyurl.com/nllbflorestranslations}{tinyurl.com/nllbflorestranslations}} For populating the post-editing prompt, we rely on glossary entries from \textsc{Gatitos}, and on parallel sentences from \href{https://huggingface.co/datasets/allenai/nllb}{allenai/nllb}, which is based on the NLLB data mining strategy.

\paragraph{Models}
We compare MT performance using ChrF++ (given the lack of neural metrics available for the languages we work with) on the following setups: (1) MT only, using NLLB-54B (2) LLM only with 10 fixed examples (3) MT + LLM APE (our solution). We use Gemini 2.0 Flash \cite{gemini_team_introducing_2024} as LLM throughout our experiments.

\begin{table*}
\begin{tabularx}{\linewidth}{XSSSSSSS}
\toprule
Method & {\textsc{tpi}} & {\textsc{dzo}} & {\textsc{quy}} & {\textsc{run}} & {\textsc{lin}} & {\textsc{asm}} & {\textsc{\textbf{average}}} \\
\midrule
\textit{Baselines: NMT / LLM only} & & & & & & \\
NLLB 54B & 41.61 & 34.67 & 26.87 & 42.51 & 47.99 & 35.91 & 38.26 \\
Gemini, 10-shot & 44.07 & 30.74 & 31.28 & 39.83 & 49.42 & 38.29 & 38.94 \\
\midrule
\textsc{TULUN}: \textit{NMT + LLM APE} & & & & & & \\
NLLB + Gemini APE & 46.80 & 35.76 & 32.40 & 41.17 & 50.58 & 39.80 & 41.09 \\
\quad $\Delta$ vs NLLB 54B & \highposdelta{+5.19} & \posdelta{+1.09} & \highposdelta{+5.53} & \negdelta{-1.34} & \posdelta{+2.59} & \highposdelta{+3.89} & \posdelta{+2.83} \\
\bottomrule
\end{tabularx}
\caption{ChrF++ score comparison on FLORES for 6 low-resource languages, using the Gatitos glossary and sentences from the NLLB training set in the translation memory.}
\label{tab:experiments}
\end{table*}

\paragraph{Results}
Our system achieves higher average accuracy than both baselines, by 2.83 and 2.15 ChrF++ points for NLLB and Gemini respectively (Table~\ref{tab:experiments}). Interestingly, we find that Gemini often beats NLLB-54, but that our system tends to improve on NLLB or Gemini, whichever is higher. One exception, Rundi (-1.34 points), is discussed in Section~\ref{sec:discussion}.

This evaluation shows the effectiveness of our approach, even on general domain benchmarks like FLORES-200. The sharp difference in accuracy gains between this experiment and the specialized domain evaluation in Section~\ref{sec:applied-scenarios} shows that our system is most useful for specialized domains, where adaptation to new terminology and translation style is needed most.

\section{Discussion}
\label{sec:discussion}

\paragraph{Accuracy Across Languages}
While our solution is effective in both applied and theoretical scenarios, the impact of LLM post-editing on MT accuracy varies (Table~\ref{tab:experiments}), including a negative effect for Rundi (-1.34 ChrF++ points). Through qualitative analysis and evaluation without injecting the glossary in the prompt for Rundi (resulting in +0.25 points compared to NLLB), we find this is due to incorrect word changes by the LLM using the glossary, highlighting the need for prompt tuning, and for glossary adjustments.

\paragraph{User-friendliness and Adaptability}
Our usability study (\cref{sec:human-eval}) confirms \textsc{Tulun}'s ease of use, but the system's configurability (\cref{sec:sys-config}) presents a potential trade-off: while it allows users to adapt the system to their needs, it also requires some understanding of MT and LLM options. Future work could explore intelligent defaults and guidance to improve accessibility, including a system module for prompt tuning (see also \cref{para:prompt-tuning-future-work}).

\paragraph{Explainability}
The transparency provided by displaying glossary matches and translation memory hits helps users understand how the system post-edited translations (see response to question 5 in Appendix~\ref{app:usability-usefulness}), but relies on the LLM's capability to use these resources effectively. For extremely low-resource languages with complex morphology or rare scripts, where LLMs have minimal prior language exposure, this assumption might not hold, resulting in higher rates of hallucination.

\section{Conclusion \& Future Work}

In this work, we present \textsc{Tulun} an open-source translation system that combines MT with LLM-based post-editing for a more accurate and adaptable low-resource translation. By leveraging existing glossaries and translation memories to guide the post-editing process, our approach achieves significant improvements over standalone MT, without requiring model fine-tuning or technical expertise. It also introduces a change of paradigm in MT, where end-users are given the opportunity to constantly improve the translation process, fostering a transparent, collaborative process that respects local expertise. Reflecting on our design and development of this system, we lay out the following future research directions:

\paragraph{Prompt Engineering and Optimization}
\label{para:prompt-tuning-future-work}
While our current prompt design yields promising results, future work could explore systematic prompt engineering approaches to maximize post-editing accuracy. This includes automatically generating language-specific prompts using techniques like DSPy's MIPRO \cite{khattab_dspy_2024}, optimizing few-shot examples based on error patterns, and developing prompts that better handle linguistic nuances in different target languages.

\paragraph{Offline Deployment Option}
To better serve users with limited internet connectivity, we plan to explore lightweight LLM options that can run locally. This likely would involve specialized small models fine-tuned specifically for the post-editing task, enabling organizations to maintain terminology consistency without relying on cloud-based LLM providers.



\section*{Ethics and Broader Impact Statement}

\textsc{Tulun} is designed to augment human translation expertise rather than replace it, particularly for low-resource languages where professional translation resources are limited. The Tetun medical glossary and Bislama dictionary used in our evaluations belong to their respective organizations and were used with explicit permission for research purposes. Usability study participants engaged voluntarily in this research and have been actively using the system since its creation. Two of the participants are co-authors of this paper, ensuring their contributions are properly acknowledged and used directly to inform our system design and evaluation.

We recognize that translation technologies can impact professional translators' workflows, and \textsc{Tulun}'s interface aims to give users control over the translation process while maintaining human oversight, especially for sensitive domains like health. We acknowledge that the system's effectiveness will vary across languages and domains, and plan to further research language-specific limitations that warrant refinement.

\bibliography{custom, anthology_1, zotero}

\appendix

\section{Example Prompt for Post-editing}
\label{sec:example-prompt}

\newtcolorbox{pretransbox}{
  colback=gray!10,        
  boxrule=0.5pt,          
  arc=0pt,                
  outer arc=0pt,
  left=10pt,              
  right=10pt,             
  top=3pt,                
  bottom=3pt,             
  bottomrule=0pt,       
  toprule=0.5pt,          
  leftrule=0pt,           
  rightrule=0pt,          
  width=\linewidth,       
}

\begin{pretransbox}
\textsc{System}
\end{pretransbox}

\noindent
You are an expert translator. I am going to give you relevant glossary entries, and relevant past translations, where the first is the English source, the second is a machine translation of the English to Tetun, and the third is the Tetun reference translation. The sentences will be written

\noindent English: <sentence>

\noindent MT: <machine translated sentence>

\noindent Tetun: <translated sentence>.

\vspace{3mm}

After the example pairs, I am going to provide another sentence in English and its machine translation, and I want you to translate it into Tetun. Give only the translation, and no extra commentary, formatting, or chattiness. Translate the text from English to Tetun.

\begin{pretransbox}
\textsc{User}
\end{pretransbox}

\noindent <glossary entries>

\noindent no 0: check -> vt. kontrola.

\noindent no 1: burn -> n. keimadura (ahi-haan)

\noindent no 2: assessment -> n. avaliasaun.

\noindent </glossary entries>

\vspace{3mm}

\noindent <past translations>

\noindent English: Antibiotic prophylaxis for burns, wounds and bites, and treatment

\noindent MT: Profilaxia antibiótiku ba kanek, feridu no morde, no tratamentu

\noindent Tetun: Ba profilaxia antibiotiku kelmadura (ai-han), kanek, tata, tohar (tohar nakloke), no tratamentu.

\noindent ...

\noindent 

\noindent </past translations>

\vspace{3mm}

\noindent Text to translate:

\noindent English: Always check burn again a couple of hours after first assessment, unless burn has been dressed.

\noindent MT: Sempre kontrola tan kanek rua oras hafoin avaliasaun dahuluk, la’ós kanek ne’ebé hetan tratamentu

\noindent Tetun: 

\begin{pretransbox}
\textsc{Assistant}
\end{pretransbox}

\noindent Sempre kontrola fali keimadura (ahi-haan) iha oras balun nia laran depois de avaliasaun dahuluk, se karik keimadura falun ona.

\tcbset{
    promptbox/.style={
        colback=white,
        colframe=gray!70!black,
        fonttitle=\color{white},
        parbox=false,
        halign=left,
        top=0.5em,
        bottom=0.5em,
        left=0.5em,
        right=0.5em,
    }
}

\lstset{
    breaklines=true,
    breakatwhitespace=false,
    columns=flexible,
    basicstyle=\normalsize\ttfamily,
    backgroundcolor=\color{gray!20},
    keepspaces=true
}

\newpage

\section{Usability and Usefulness Responses}
\label{app:usability-usefulness}

\begin{table}[h]
\centering
\small
\begin{tabularx}{\linewidth}{XSS}
\toprule
Statement & {R1} & {R2} \\
\midrule
1. I think that I would like to use this system frequently. & 5 & 5 \\
2. I found the system unnecessarily complex. & 1 & 1 \\
3. I thought the system was easy to use. & 5 & 5 \\
4. I think that I would need the support of a technical person to be able to use this system. & 1 & 2 \\
5. I found the various functions in this system were well integrated. & 1 & 2 \\
6. I thought there was too much inconsistency in this system. & 1 & 2 \\
7. I would imagine that most people would learn to use this system very quickly. & 5 & 3 \\
8. I found the system very cumbersome to use. & 1 & 2 \\
9. I felt very confident using the system. & 5 & 4 \\
10. I needed to learn a lot of things before I could get going with this system. & 2 & 2 \\
\bottomrule
\end{tabularx}
\caption{Usability ratings}
\label{tab:usability}
\end{table}

\begin{table}[h]
\centering
\small
\begin{tabularx}{\linewidth}{XSS}
\toprule
Statement & {R1} & {R2} \\
\midrule
1. Using this system improves the quality of my translations. & 5 & 4 \\
2. Using this system increases my productivity when translating documents. & 5 & 4 \\
3. Using this system enhances my effectiveness in maintaining terminology consistency. & 4 & 4 \\
4. Using this system makes it easier to translate technical/specialized content. & 4 & 5 \\
5. The glossary and translation memory features are useful for my translation work. & 5 & 5 \\
6. Overall, I find this system useful for my translation tasks. & 5 & 5 \\
\bottomrule
\end{tabularx}
\caption{Usefulness ratings}
\label{tab:usefulness}
\end{table}

\end{document}